\def\year{2020}\relax
\documentclass[letterpaper]{article} % DO NOT CHANGE THIS
\usepackage{aaai20}  % DO NOT CHANGE THIS
\usepackage{times}  % DO NOT CHANGE THIS
\usepackage{helvet} % DO NOT CHANGE THIS
\usepackage{courier}  % DO NOT CHANGE THIS
\usepackage[hyphens]{url}  % DO NOT CHANGE THIS
\usepackage{graphicx} % DO NOT CHANGE THIS
\usepackage{graphicx}  % DO NOT CHANGE THIS
\usepackage{amsmath}
\usepackage{booktabs}

\title{Back to the Future -- Sequential Alignment of Text Representations}

\author{Johannes Bjerva,\textsuperscript{\rm 1*} Wouter M. Kouw,\textsuperscript{\rm 1,2}\thanks{JB and WMK contributed equally to this work.}
Isabelle Augenstein\textsuperscript{\rm 1}\\  % All authors must be in the same font size and format. Use \Large and \textbf to achieve this result when breaking a line
\textsuperscript{\rm 1}Department of Computer Science, University of Copenhagen, Copenhagen, Denmark \\ 
\textsuperscript{\rm 2}Department of Electrical Engineering, Eindhoven University of Technology, Eindhoven, the Netherlands\\
bjerva@di.ku.dk, w.m.kouw@tue.nl, augenstein@di.ku.dk
}

\begin{document}

\maketitle

\begin{abstract}
  Language evolves over time in many ways relevant to natural language processing tasks. For example, recent occurrences of tokens 'BERT' and 'ELMO' in publications refer to neural network architectures rather than persons. This type of temporal signal is typically overlooked, but is important if one aims to deploy a machine learning model over an extended period of time. In particular, language evolution causes data drift between time-steps in sequential decision-making tasks. Examples of such tasks include prediction of paper acceptance for yearly conferences (regular intervals) or author stance prediction for rumours on Twitter (irregular intervals). Inspired by successes in computer vision, we tackle data drift by sequentially aligning learned representations. %We consider both unsupervised and semi-supervised alignment. 
  We evaluate on three challenging tasks varying in terms of time-scales, linguistic units, and domains. These tasks show our method outperforming several strong baselines, including using all available data. We argue that, due to its low computational expense, sequential alignment is a practical solution to dealing with language evolution.
  %Furthermore, we show that this alignment can be learned from as few as two future data points.
%  These results support the importance of temporal annotation of NLP data and exploitation thereof.%, particularly as this important signal can often be obtained nearly for free.
\end{abstract}

\section{Introduction}
\noindent As time passes, language usage changes.
% These changes occur both in noticeable ways, for instance when new words are coined or when existing words are used differently; and in less noticeable ways when, e.g., through changes in grammar.
For example, the names `Bert' and `Elmo' would only rarely make an appearance prior to 2018 in the context of scientific writing. After the publication of BERT \cite{bert} and ELMo \cite{elmo}, however, usage has increased in frequency. In the context of named entities on Twitter, it is also likely that these names would be tagged as \textsc{person} prior to 2018, and are now more likely to refer to an \textsc{artefact}. As such, their part-of-speech tags will also differ. Evidently, evolution of language usage affects natural language processing (NLP) tasks, and as such, models based on data from one point in time cannot be expected to generalise to the future.

In order to become more robust to language evolution, data should be collected at multiple points in time. We consider a dynamic learning paradigm where one makes predictions for data points from the current time-step given labelled data points from previous time-steps. As time increments, data points from the current step are labelled and new unlabelled data points are observed. This setting occurs in NLP in, for instance, the prediction of paper acceptance to conferences \cite{kang18naacl} or named entity recognition from yearly data dumps of Twitter \cite{derczynski:2016}. Changes in language usage cause a data drift between time-steps and some way of controlling for the shift between time-steps is necessary. 
% In the above two examples, time progresses in regular intervals (i.e. yearly), but there are tasks where time progresses at irregular intervals as well. For example, author stance recognition for rumours on Twitter \cite{rumour:19}. In the RumourEval 2019 data set, time between rumours can vary between less than a month and more than a year. These cases are arguably more difficult as there is more uncertainty on how much the data has drifted. 

%Based on successes in image processing, and inspired by the successes of alignment methods in other NLP tasks \cite{lample:2018}, we experiment with \textbf{temporal domain adaptation} (TDA), based on semi-supervised subspace alignment, and show that this method yields substantial improvements.

In this paper, we apply a domain adaptation technique to correct for shifts. Domain adaptation is a furtive area of research within machine learning that deals with learning from training data drawn from one data-generating distribution (source domain) and generalising to test data drawn from another, different data-generating distribution (target domain) \cite{kouw2019review}. %In practice, one adapts from a labelled data set to an unlabelled data set. 
We are interested in whether a sequence of adaptations can compensate for the data drift caused by shifts in the meaning of words or features across time. Given that linguistic tokens are embedded in some vector space using neural language models, we observe that in time-varying dynamic tasks, the drift causes token embeddings to occupy different parts of embedding space over consecutive time-steps. We want to avoid the computational expense of re-training a neural network every time-step. Instead, in each time-step, we map linguistic tokens using the same pre-trained language model (a "BERT" network \cite{bert}) and align the resulting embeddings using a second procedure called subspace alignment \cite{fernando2013unsupervised}. We apply subspace alignment sequentially: find the principal components in each time-step and linearly transform the components from the previous step to match the current step. A classifier trained on the aligned embeddings from the previous step will be more suited to classify embeddings in the current step. We show that sequential subspace alignment (SSA) yields substantial improvements in three challenging tasks: paper acceptance prediction on the PeerRead data set \cite{kang18naacl}; Named Entity Recognition on the Broad Twitter Corpus \cite{derczynski:2016}; and rumour stance detection on the RumourEval 2019 data set \cite{rumour:19}. %RumourEval-17 and 19 data sets \cite{rumour:17}.
These tasks are chosen to vary in terms of domains, timescales, and the granularity of the linguistic units. %, while containing timestamps required for SSA. %in addition to the restriction that we require data with timestamps
%(Table~\ref{tab:quickdata}).
In addition to evaluating SSA, we include two technical contributions as we extend the method both to allow for time series of unbounded length and to consider instance similarities between classes.
The best-performing SSA methods proposed here are semi-supervised, but require only between 2 and 10 annotated data points per class from the test year for successful alignment.
Crucially, the best proposed SSA models outperform baselines utilising more data, including the whole data set. 
%In all cases TDA improves upon strong baselines, including the setting of having access to \textit{all} training data, including future data. We further show that, depending on the task at hand, access to between 2 and 10 annotated data points per class from the test year can significantly improve performance when using semi-supervised TDA.

%The paper is outlined as follows: the next section presents subspace alignment including our extensions. Following that, we present our experiments and results. Afterwards, we show an example of alignment applied to tweets, a discussion of the limitations of our method and connections to related work. 

\section{Subspace Alignment}
\begin{figure}[!t]
    \centering
    \includegraphics[width=.44\textwidth]{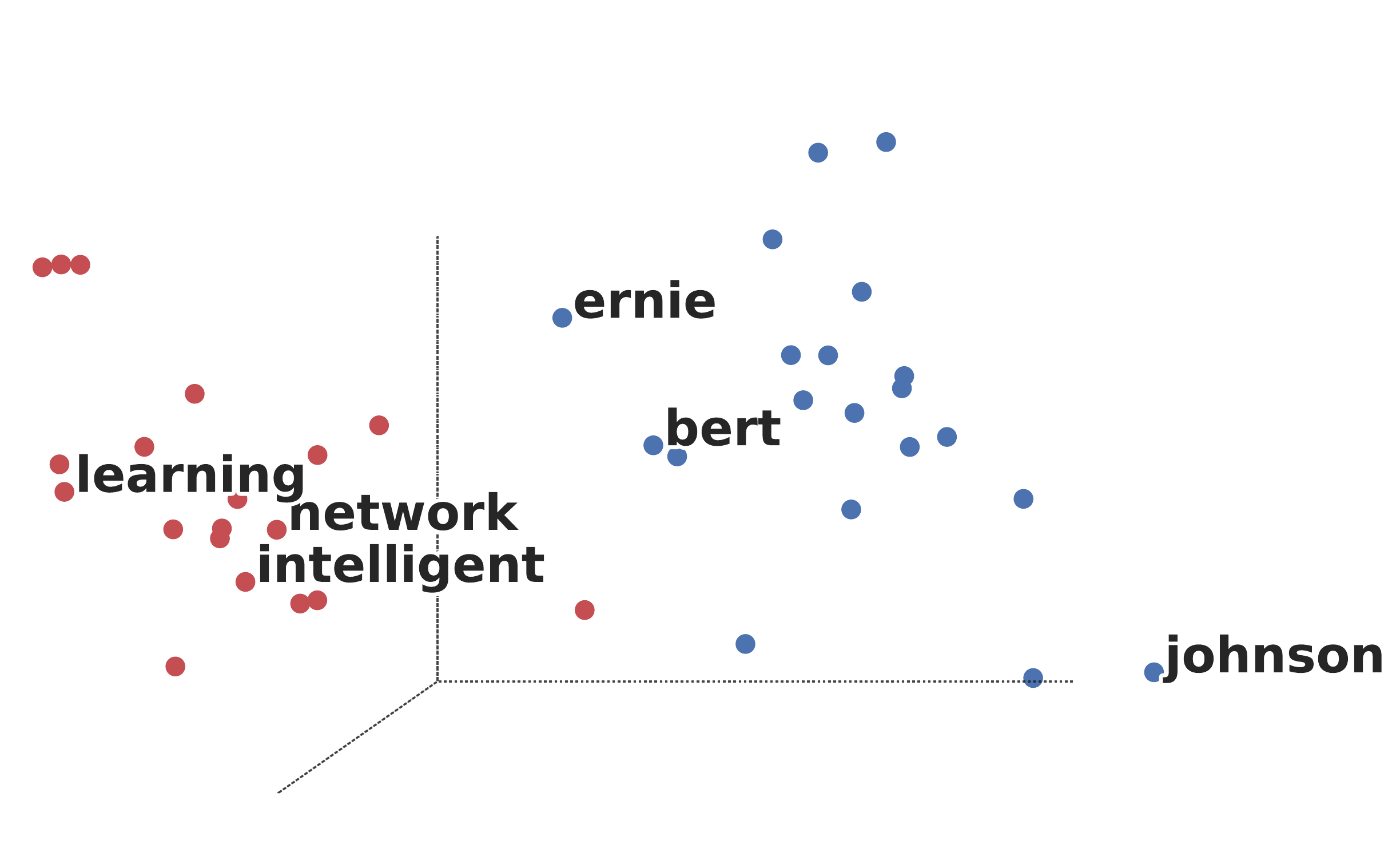}
    \includegraphics[width=.44\textwidth]{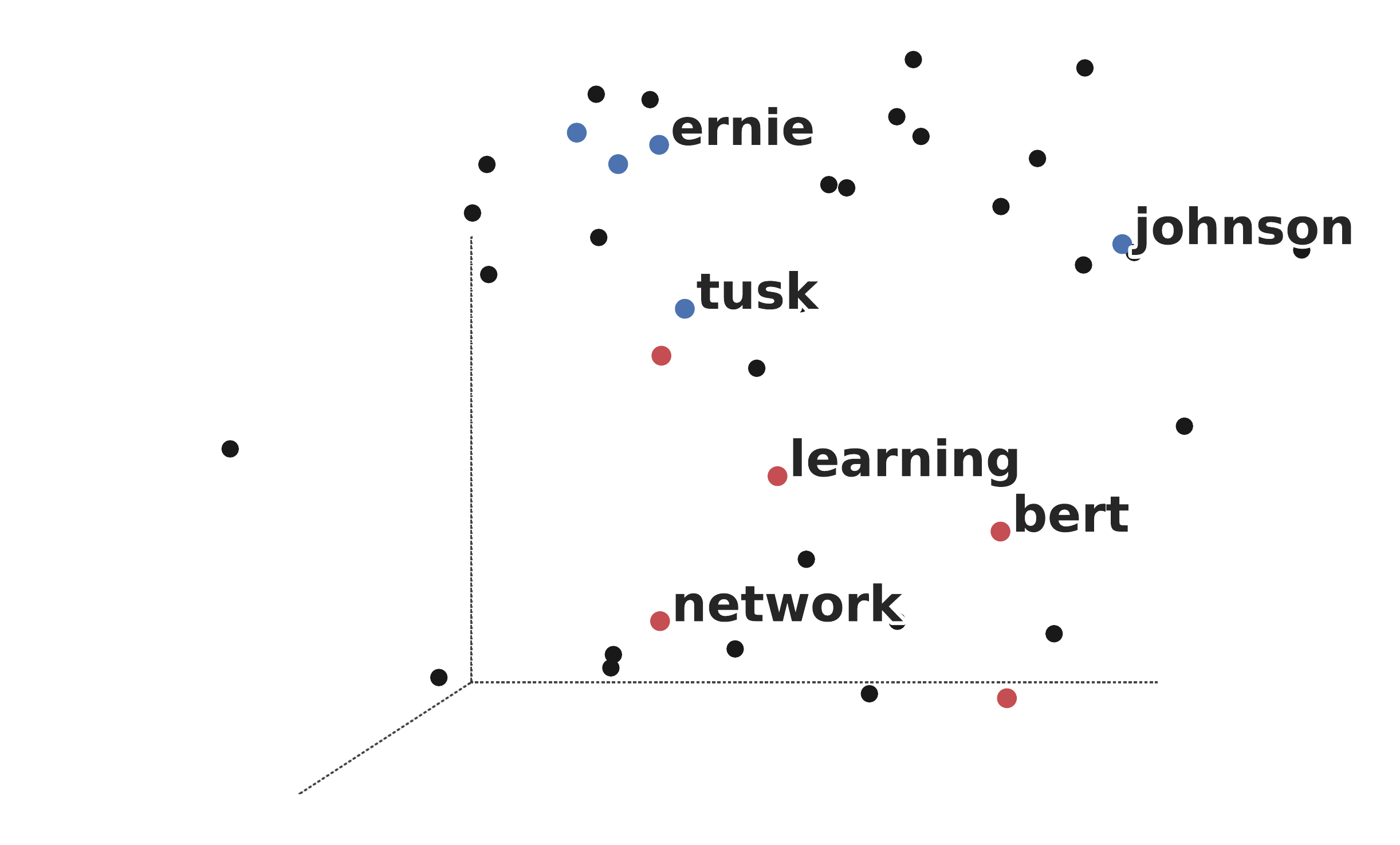}
    \caption{Example of a word embedding at $t_{2017}$ vs $t_{2018}$ (blue={\sc person}, red={\sc artefact}, black={\sc unk}). Source data (top, $t_{2017}$), target data (bottom, $t_{2018}$). Note that at $t_{2017}$, 'bert' is a {\sc person}, while at $t_{2018}$, 'bert' is an {\sc artefact}.}
    \label{fig:2DG}
\end{figure}

Suppose we embed words from a named entity recognition task, where {\sc artefact}s should be distinguished from {\sc person}s. Figure \ref{fig:2DG} shows scatterplots with data collected at two different time-points, say 2017 (top; source domain) and 2018 (bottom; target domain). Red points are examples of {\sc artefact}s  embedded in this space and blue points are examples of {\sc person}s. We wish to classify the unknown points (black) from 2018 using the labeled points (red/blue bottom) from 2018 and the labeled points from 2017 (red/blue top).

As can be seen, the data from 2017 is not particularly relevant to classification of data from 2018, because the red and blue point clouds do not match. In other words, a classifier trained to discriminate red from blue in 2017 would make a lot of mistakes when applied directly to the data from 2018, partly because words such as 'bert' have changed from being {\sc person}s to being {\sc artefact}s. %We will refer to the data from 2017 as source data and that of 2018 as target data. 
To make the source data from 2017 relevant -- and reap the benefits of having more data -- we wish to \emph{align} source and target data points.

\begin{figure*}[!t]
    \centering
    \includegraphics[width=0.95\textwidth]{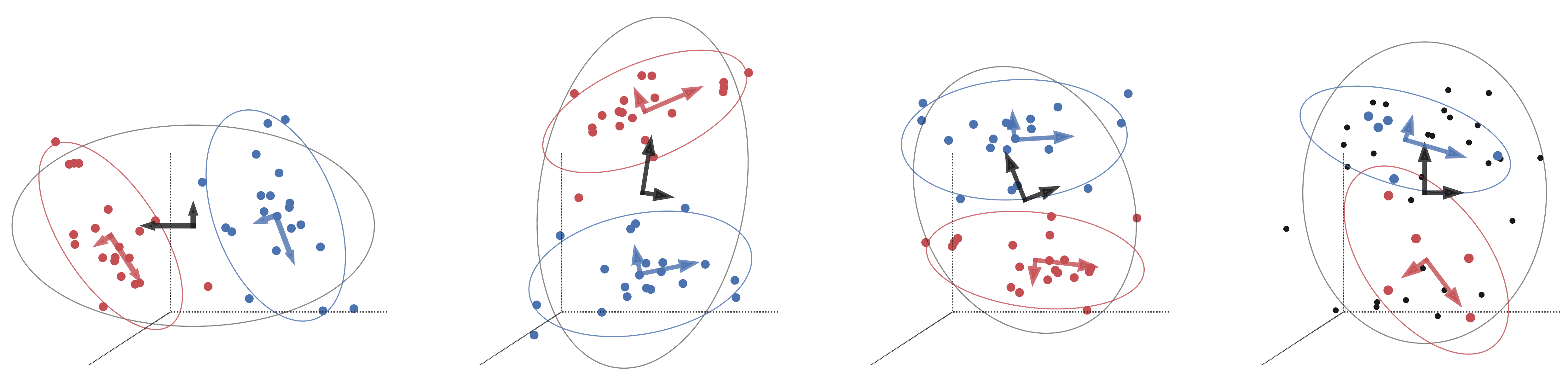}
    \caption{Illustration of subspace alignment procedures. Red vs blue dots indicate samples from different classes, arrows (black for total data and red vs blue for each class) indicate scaled eigenvectors of the covariance matrix (error ellipses indicate regions within 2 standard deviations). (Leftmost) Source data, fully labeled. (Left middle). Unsupervised subspace alignment: the total principal components from the source data (black arrows in leftmost figure) have been aligned to the total principal components of the target data (black arrows in rightmost figure). (Right middle) Semi-supervised subspace alignment: the class-specific principal components of the source data (red/blue arrows from leftmost figure) have been aligned to the class-specific components of the target data (red/blue arrows from the rightmost figure). Note that unsupervised alignment fails to match the red and blue classes across domains, while semi-supervised alignment succeeds. (Rightmost) Target data, with few labeled samples per class (black dots are unlabeled samples).}
    \label{fig:2DG_subalign}
\end{figure*}

\subsection{Unsupervised Subspace Alignment}
Unsupervised alignment extracts a set of bases from each data set and transforms the source components such that they match the target components \cite{fernando2013unsupervised}. Let $C_{\cal S}$ be the principal components of the source data $X_{t-1}$ and $C_{\cal T}$ be the components of the target data set $X_t$. The optimal linear transformation matrix is found by minimising the difference between the transformed source components and the target components:
\begin{align}
    M^{*} =& \ \underset{M}{\arg \min} \ \| C_{\cal S}M - C_{\cal T}\|^{2}_{F} \nonumber \\
    =& \ \underset{M}{\arg \min} \ \| C_{\cal S}^{\top} C_{\cal S}M - C_{\cal S}^{\top}C_{\cal T}\|^{2}_{F} \nonumber \\
    =& \ \underset{M}{\arg \min} \ \| M - C_{\cal S}^{\top}C_{\cal T}\|^{2}_{F} \ = \ C_{\cal S}^{\top} C_{\cal T} \label{eq:unsup_align} \, ,
\end{align}
where $\| \cdot \|_{F}$ denotes the Frobenius norm. Note that we left-multiplied both terms in the norm with the same matrix $C_{\cal S}^{\top}$ and that due to orthonormality of the principal components, $C_{\cal S}^{\top} C_{\cal S}$ is the identity and drops out. 
Source data $X_{t-1}$ is aligned to target data by first mapping it onto its own principal components and then applying the transformation matrix, $X_{t-1} C_{\cal S} M^{*}$. Target data $X_t$ is also projected onto its target components, $X_t C_{\cal T}$. 
The alignment is performed on the $d$ largest principal components, i.e.~a \emph{subspace} of the embedding. Keeping $d$ small avoids the high computational expense of eigendecomposition in high-dimensional data. 

Unsupervised alignment will only match the total structure of both data sets. Therefore, global shifts between domains can be accounted for, but not local shifts. Figure \ref{fig:2DG} is an example of a setting with local shifts, i.e. red and blue classes are shifted differently. Performing unsupervised alignment on this setting would fail. Figure  \ref{fig:2DG_subalign} (left middle) shows the source data (leftmost) aligned to the target data (rightmost) in an unsupervised fashion. Note that although the total data sets roughly match, the classes (red and blue ellipses) are not matched.

\subsection{Semi-Supervised Subspace Alignment}
In semi-supervised alignment, one performs subspace alignment \textit{per class}. As such, at least $1$ target label per class needs to be available. However, even then, with only $1$ target label per class, we would only be able to find $1$ principal component. To allow for the estimation of more components, we provisionally label all target samples using a $1$-nearest-neighbour classifier, starting from the given target labels. Using pseudo-labelled target samples, we estimate $d$ components.

Now, the optimal linear transformation matrix for each class can be found with an equivalent procedure as in Equation \ref{eq:unsup_align}:
\begin{align}
    M_{k}^{*} = \ \underset{M}{\arg \min} \ \| C_{{\cal S}, k} M - C_{{\cal T},_k} \|^{2}_{F} \ = \ C_{{\cal S},k}^{\top} C_{{\cal T},k} \label{eq:semisup_align} \, .
\end{align}
Afterwards, we transform the source samples of each class $X_{t-1}^k$ through the projection onto class-specific components $C_{{\cal S},k}$ and the optimal transformation: $X_{t-1}^{k} C_{{\cal S}, k} M_{k}^{*}$. Additionally, we centre each transformed source class on the corresponding target class. Figure \ref{fig:2DG_subalign} (right middle) shows the source documents transformed through semi-supervised alignment. Now, the classes match the target data classes.
% Note that in Figure \ref{fig:2DG} the  are opposite of each other, but that now - after transformation - they match nicely.

%%% Entire assumption is in Limitations section
%  We make the explicit assumption that the embedded documents are clustered by class, i.e., that most red points are more similar to other red points than to blue points. That means that, in the target domain, we can assume that the unlabelled documents have a high probability of belonging to the same class as their closest labelled documents. In turn, this implies that when we align the red cluster in the source domain to the red points in the target domain, we assume that the closest black points belong to the red class as well. %We discuss an interpretation of this assumption in Section \ref{sec:discussion}.

\subsection{Extending SSA to Unbounded Time}
Semi-supervised alignment allows for aligning \emph{two} time steps, $t_1$ and $t_2$, to a joint space $t'_{1,2}$.
However, when considering a further alignment to another time step $t_3$, this can not trivially be mapped, since the joint space $t'_{1,2}$ necessarily has a lower dimensionality.
Observing that two independently aligned spaces, $t'_{1,2}$ and $t'_{2,3}$, \emph{do} have the same dimensionality, we further learn a new alignment between the two, resulting in the joint space of $t'_{1,2}$ and $t'_{2,3}$, namely $t''_{1,2,3}$.
While there are many ways of joining individual time steps to a single joint space, we approach this by building a binary branching tree, first joining adjacent timesteps with each other, and then joining the new adjacent subspaces with each other.

Although this is seemingly straight-forward, there is no guarantee that $t'_{1,2}$ and $t'_{2,3}$ will be coherent with one another, in the same way that two word embedding spaces trained with different algorithms might also differ in spite of having the same dimensionality.
This issue is partially taken care of by using semi-supervised alignment which takes class labels into account when learning the 'deeper' alignment $t''$.
We further find that it is beneficial to also take the similarities between samples into account when aligning.

\subsection{Considering Sample Similarities between Classes}

Since intermediary spaces, such as  $t'_{1,2}$ and $t'_{2,3}$, do not necessarily share the same semantic properties, we add a step to the semi-supervised alignment procedure.
Given that the initial unaligned spaces do encode similarities between instances, we run the $k$-means clustering algorithm ($k=5$) to give us some course-grained indication of instance similarities in the original embedding space.
This cluster ID is passed to SSA, resulting in an alignment which both attempts to match classes across time steps, in addition to instance similarities.
Hence, even though $t'_{1,2}$ and $t'_{2,3}$ are not necessarily semantically coherent, an alignment to $t''_{1,2,3}$ is made possible.

\section{Experimental Setup} \label{sec:experiments}
In the past year, several approaches to pre-training representations on language modelling based on transformer architectures \cite{vaswani2017} have been proposed.
These models essentially use a multi-head self-attention mechanism in order to learn representations which are able to attend directly to any part of a sequence. % -- this is in contrast with an LSTM, which are also able to learn this type of focus implicitly, but are forced to pass this type of signal throughout an entire linear sequence.
Recent work has shown that such contextualised representations pre-trained on language modelling tasks offer highly versatile representations which can be fine-tuned on seemingly any given task \cite{elmo,bert,gpt,gpt2}.
% Maybe for camera ready:
%In addition to the empirical results this leads to, we see this as a substantial experimental advantage.
%In essence, having this type of pre-training step allows for constructing cleaner NLP experiments in which the representations used up to a high level are identical across tasks. \johannes{I think it's nice to try to make the philosophical point here that BERT etc. allows for cleaner experimentation.} 
%This can be compared with previous approaches of pre-training word embeddings, potentially with language modelling-resembling tasks (e.g., \citeauthor{mikolov:2013}), which at most could yield a task-agnostic representation at the word level, which in turn would have a large number of task-specific parameters stacked on top in the form of, e.g., an $n$ layer deep LSTM \cite{lstm}.
%\johannes{term for TDA buckets}
In line with the recommendations from experiments on fine-tuning representations \cite{ruder:freeze}, we use a frozen BERT to extract a consistent task-agnostic representation. Using a frozen BERT with subsequent subspace alignment allows us to avoid re-training a neural network each time-step while still working in an embedding learned by a neural language model. It also allows us to test the effectiveness of SSA without the confounding influence of representation updates.

\paragraph{Three Tasks.}
We consider three tasks representing a broad selection of natural language understanding scenarios: paper acceptance prediction based on the PeerRead data set \cite{kang18naacl}, Named Entity Recognition (NER) based on the Broad Twitter Corpus \cite{derczynski:2016}, and author stance prediction based on the RumEval-19 data set \cite{rumour:19}.
These tasks were chosen so as to represent i) different textual domains, across ii) differing time scales, and iii) operating at varying levels of linguistic granularity.
As we are dealing with dynamical learning, the vast majority of NLP data sets can unfortunately not be used since they do not include time stamps.

\section{Paper Acceptance Prediction}
\label{sec:peerread}

\begin{table*}[!t]
    \centering
\resizebox{1.8\columnwidth}{!}{
    \begin{tabular}{lrrr|rr|rrr}
    \toprule
    \textbf{Test year} & \textbf{All} & \textbf{Same} & \textbf{Prev} & \textbf{Unsup.} & \textbf{Semi-sup.} & \textbf{Unsup. Unb.} & \textbf{S. Unb.} & \textbf{S. Unb. w/Clst} \\
    \midrule
    2010      & 61.77   & 67.64 & 35.29 & \textbf{70.59} & \textbf{70.59} & 70.58 &  \textbf{70.59}  & \textbf{70.59} \\
    2011      & 61.77   & 58.82 & 55.88 & 14.71          & \textbf{72.35} & 24.71 &  \textbf{72.35}  & \textbf{72.35} \\
    2012      & 56.25   & 56.25 & 58.75 & 50.00          & 72.50 & 45.00 &  \textbf{72.80}  & 72.30 \\
    2013      & 67.54   & 56.14 & 58.78 & 76.31          & 78.07 & 72.31 &  78.97  & \textbf{79.03} \\
    2014      & 50.53   & 51.64 & 51.64 & 36.88          & 68.03 & 31.88 &  69.03  & \textbf{69.45}      \\
    2015      & 57.83   & 54.05 & 54.05 & 49.19          & 58.37 & 41.19 &  \textbf{59.97}  & 59.93      \\
    2016      & 58.89   & 57.36 & 57.36 & 50.61          & 61.04 & 38.61 &  \textbf{63.04}  & \textbf{63.04}      \\
    2017      & 56.04   & 58.24 & 58.24 & 68.13 & 63.73          & 58.13 &  68.73  & \textbf{69.80}      \\
    \midrule
    avg       & 58.82   & 57.52 & 53.75 & 52.05          & 68.09 & 47.80 & 69.44 & \textbf{69.56} \\
    \bottomrule
    \end{tabular}
    }
    \caption{Paper acceptance prediction (acc.) on the PeerRead data set \cite{kang18naacl}. Abbreviations represent Unsupervised, Semi-supervised, Unsupervised Unbounded, Semi-supervised Unbounded, and Semi-supervised Unbounded with Clustering.}
    \label{tab:acceptance_bert}
\end{table*}

The PeerRead data set contains papers from ten years of arXiv history, as well as papers and reviews from major AI and NLP conferences \cite{kang18naacl}.\footnote{\url{https://github.com/allenai/PeerRead}} 
From the perspective of evaluating our method, the arXiv sub-set of this data set offers the possibility of evaluating our method while adapting to ten years of history.
This is furthermore the only subset of the data annotated with both timestamps and with a relatively balanced accept/reject annotation.\footnote{The NIPS selection, ranging from 2013-2017, only contains accepted papers. The other conferences contain accept/reject annotation, but only represent single years.}
As arXiv naturally contains both accepted and rejected papers, this acceptance status has been assigned based on \citeauthor{sutton:2017} who match arXiv submissions to bibliographic entries in DBLP, and additionally defining acceptance as having been accepted to major conferences, and not to workshops.
This results in a data set of nearly 12,000 papers, from which we use the raw abstract text as input to our system. The first three years were filtered out due to containing very few papers.
We use the standard train/test splits supplied with the data set.

\citeauthor{kang18naacl} show that it is possible to predict paper acceptance status at major conferences at above baseline levels.
Our intuition in applying SSA to this problem, is that the topic of a paper is likely to bias acceptance to certain conferences \textit{across time}.
For instance, it is plausible that the likelihood of a neural paper being accepted to an NLP conference before and after 2013 differs wildly.
Hence, we expect that our model will, to some extent, represent the topic of an article, and that this will lend itself nicely to SSA.

\subsection{Model}
\begin{figure}[!t]
    \centering
    \includegraphics[width=0.75\columnwidth]{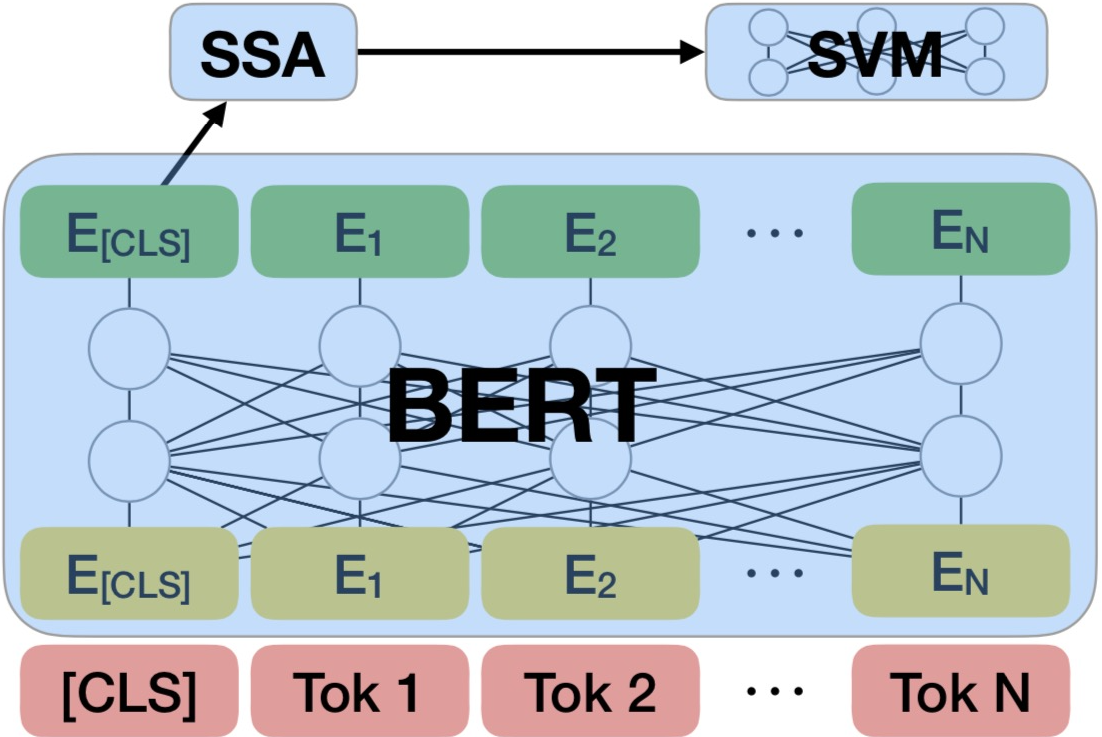}
    \caption{Paper acceptance model (BERT and SSA).}
    \label{fig:bertseq}
\end{figure}

We use the pre-trained \textsc{bert-base-uncased} model as the base for our paper acceptance prediction model.
Following the approach of \citeauthor{bert}, we take the final hidden state (i.e., the output of the transformer) corresponding to the special \texttt{[CLS]} token of an input sequence to be our representation of a paper, as this has aggregated information through the sequence (Figure~\ref{fig:bertseq}).
%Concretely, if 
%\begin{equation}
%p({\boldsymbol \pi}) = \prod_{i=1}^{|\bold{\pi}|} p(\pi_i \mid \textit{pa}_{{\cal T}}[\pi_i])
%\end{equation}
This gives us a $d$-dimensional representation of each document, where $d=786$.
In all of the experiments for this task, we train an SVM with an RBF kernel on these representations, either with or without SSA depending on the setting.

\subsection{Experiments \& Results}
We set up a series of experiments where we observe past data, and evaluate on present data. 
We compare both unsupervised and semi-supervised subspace alignment, with several strong baselines.
The baselines represent cases in which we have access to more data, and consist of training our model on either \textbf{all} data (i.e.~both past and future data), on the \textbf{same} year as the evaluation year, and on the \textbf{previous} year.
In our alignment settings, we only observe data from the previous year, and apply subspace alignment.
This is a different task than presented by \citeauthor{kang18naacl}, as we evaluate paper acceptance for papers in the present. Hence, our scores are not directly comparable to theirs.

\begin{figure}[!t]
    \centering
    \includegraphics[width=0.8\columnwidth]{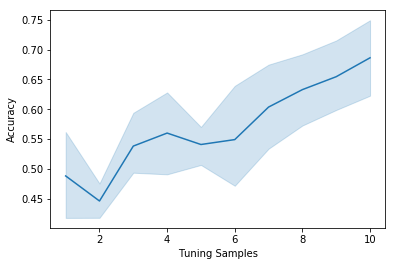}
    \caption{Tuning semi-supervised subspace alignment on PeerRead development data (95\% CI shaded).}
    \label{fig:hyperparam}
\end{figure}

One parameter which significantly influences performance, is the number of labelled data points we use for learning the semi-supervised subspace alignment.
We tuned this hyperparameter on the development set, finding an increasing trend. Using as few as 2 tuning points per class yielded an increase in performance in some cases (Figure~\ref{fig:hyperparam}).

Our results are shown in Table~\ref{tab:acceptance_bert}, using 10 tuning samples per class.
With unsupervised subspace alignment, we observe relatively unstable results -- in one exceptional case, namely testing on 2010, unsupervised alignment is as helpful as semi-supervised alignment.
Semi-supervised alignment, however, yields consistent improvements in performance across the board.
It is especially promising that adapting from past data outperforms training on all available data, as well as training on the actual in-domain data.
This highlights the importance of controlling for data drift due to language evolution. It shows that this signal can be taken advantage of to increase performance on present data with only a small amount of annotated data.
We further find that using several past time steps in the Unbounded condition is generally helpful, as is using instance similarities in the alignment.

\section{Named Entity Recognition}

\begin{table*}[!t]
    \centering
\resizebox{1.8\columnwidth}{!}{
    \begin{tabular}{lrrr|rr|rrr}
    \toprule
    \textbf{Test year} & \textbf{All} & \textbf{Same} & \textbf{Prev} & \textbf{Unsup.} & \textbf{Semi-sup.} & \textbf{Unsup. Unb.} & \textbf{S. Unb.} & \textbf{S. Unb. w/Clst} \\
    \midrule
    2013      & 62.95  & 42.24 & 54.16 & 42.25  & 63.82 & 42.25  & 63.82 & \textbf{63.95} \\
    2014      & 72.77  & 77.76 & 59.53 & 50.43  & 73.67 & 50.43  & 73.67 & \textbf{78.75} \\
    \midrule
    avg       & 67.86  & 60.00 & 56.85 & 46.34 & 68.75  & 46.34 & 68.75 &  \textbf{71.35} \\
    \bottomrule
    \end{tabular}
    }
    \caption{NER (F1 score) on the Broad Twitter Corpus \cite{derczynski:2016}.}
    \label{tab:NER}
\end{table*}

The Broad Twitter Corpus contains tweets annotated with named entities, collected between the years 2009 and 2014 \cite{derczynski:2016}.
However, as only a handful of tweets are collected before 2012, we focus our analysis on the final three years of this period (i.e.~two test years).
The corpus includes diverse data, annotated in part via crowdsourcing and in part by experts.
The inventory of tags in their tag scheme is relatively small, including Person, Location, and Organisation.
To the best of our knowledge no one has evaluated on this corpus either in general or per year, and so we cannot compare with previous work.

In the case of NER, we expect the adaptation step of our model to capture the fact that named entities may change their meaning across time (e.g. the example with ''Bert'' and ''BERT'' in Figure \ref{fig:2DG}).
%For instance, names such as ``Bert" and ``Elmo" might be tagged as \textsc{person} in a data set from before 2018, and are now more likely to refer to an \textsc{artefact}.
This is related to work showing temporal drift of topics \cite{wang2006topics}.

\subsection{Model}
Since casing is typically an important feature in NER, we use the pre-trained \textsc{bert-base-cased} model as our base for NER.
For each token, we extract its contextualised representation from BERT, before applying SSA.
As \citeauthor{bert} achieve state-of-the-art results without conditioning the predicted tag sequence on surrounding tags (as would be the case with a CRF, for example), we also opt for this simpler architecture.
The resulting contextualised representations are therefore passed to an MLP with a single hidden layer (200 hidden units, ReLU activation), before predicting NER tags.
We train the MLP over 5 epochs using the Adam optimiser \cite{adam}.

\subsection{Experiments \& Results}
As with previous experiments, we compare unsupervised and semi-supervised subspace alignment with baselines corresponding to using all data, data from the same year as the evaluation year, and data from the previous year.
For each year, we divide the data into 80/10/10 splits for training, development, and test.
Results on the two test years 2013 and 2014 are shown in Table~\ref{tab:NER}.
In the case of NER, we do not observe any positive results for unsupervised subspace alignment.
In the case of semi-supervised alignment, however, we find increased performance as compared to training on the previous year, and compared to training on all data.
This shows that learning an alignment from just a few data points can help the model to generalise from past data.
However, unlike our previous experiments, results are somewhat better when given access to the entire set of training data from the test year itself in the case of NER.
The fact that training on only 2013 and evaluating on the same year does not work well can be explained by the fact that the amount of data available for 2013 is only 10\% of that for 2012.
The identical results for the unbounded extension is because aligning from a single time step renders this irrelevant.

\section{SDQC Stance Classification}
The RumourEval-2019 data set consists of roughly 5500 tweets collected for 8 events surrounding well-known incidents, such as the Charlie Hebdo shooting in Paris \cite{rumour:19}.\footnote{\url{http://alt.qcri.org/semeval2019/index.php?id=tasks}} 
Since the shared task test set is not available, we split the training set into a training, dev and test part based on rumours (one rumour will be training data with a 90/10 split for development and another rumour will be the test data, with a few samples labelled). For Subtask A, tweets are annotated with stances, denoting whether it is in the category \texttt{Support}, \texttt{Deny}, \texttt{Query}, or \texttt{Comment} (SDQC). 

Each rumour only lasts a couple of days, but the total data set spans years, from August 2014 to November 2016.  We regard each rumour as a time-step and adapt from the rumour at time $t$-$1$ to the rumour at time $t$. We note that this setting is more difficult than the previous two due to the irregular time intervals. We disregard the rumour \texttt{ebola-essien} as it has too few samples per class.
% Note that this somewhat conflates TDA and normal domain adaptation.
% However, since the data set is quite small (approximately 5500 instances, across 8 events, and several days per event), doing TDA on time steps per day within each rumour is not feasible.

\subsection{Model}
For this task, we use the same modelling approach as described for paper acceptance prediction. % (Section~\ref{sec:peerread}).
This method is also suitable here, since we simply require a condensed representation of a few sentences on which to base our temporal adaptation and predictions.
In the last iteration of the task, the winning system used hand-crafted features to achieve a high performance \cite{S17-2083}. Including these would complicate SSA, so we opt for this simpler architecture instead.
We use the shorter time-scale of approximately weeks rather than years as rumours can change rapidly \cite{kwon2017rumor}.

\subsection{Experiments \& Results}
In this experiment, we start with the earliest rumour and adapt to the next rumour in time. As before, we run the following baselines: training on all available labelled data (i.e.~all previous rumours and the labelled data for the current rumour), training on the labelled data from the current rumour (designated as `same') and training on the labelled data from the previous rumour. We perform both unsupervised and semi-supervised alignment using data from the previous rumour. We label 5 samples per class for each rumour.

In this data set, there is a large class imbalance, with a large majority of \texttt{comment} tweets and few \texttt{support} or \texttt{deny} tweets. To address this, we over-sample the minority classes. Afterwards, a SVM with RBF is trained and we test on unlabelled tweets for the current rumour. 
Table~\ref{tab:SDQC} shows the performance of the baselines and the two alignment procedures. As with the previous tasks, semi-supervised alignment generally helps, except for in the \texttt{charliehebdo} rumour.

\begin{table*}[!t]
    \centering
\resizebox{1.8\columnwidth}{!}{
    \begin{tabular}{lrrr|rr|rrr}
    \toprule
       \textbf{Test year} & \textbf{All} & \textbf{Same} & \textbf{Prev} & \textbf{Unsup.} & \textbf{Semi-sup.} & \textbf{Unsup. Unb.} & \textbf{S. Unb.} & \textbf{S. Unb. w/Clst} \\
    \midrule
    \texttt{ottawashooting}     & 31.51  & 23.67  & 30.77  & 30.77  & \textbf{31.88}  & 28.37       & 30.68 & 30.88 \\
    \texttt{prince-toronto}     & 36.27  & 23.37  & 34.46  & 34.46  & \textbf{40.32}  & 31.36       & 39.12 & 39.52 \\
    \texttt{sydney-siege}       & 32.34  & 27.17  & 41.23  & 41.23  & \textbf{43.60}  & 33.23       & 43.50 & 43.54 \\
\texttt{charliehebdo}       & 38.51  & 31.67  & \textbf{35.73}  & \textbf{35.73}  & 33.76  & 33.71  & 32.70 & 32.61 \\
    \texttt{putinmissing}       & 28.33  & 22.38  & 34.53  & 34.53  & \textbf{36.11}  & 31.95       & 35.10 & 35.81 \\
    \texttt{germanwings-crash}  & 29.38  & 22.01  & 44.79  & 44.79  & \textbf{44.84}  & 40.30       & 44.88 & 44.80 \\
    \texttt{illary}             & 29.24  & 25.81  & 37.53  & 37.53 & \textbf{40.08}   & 34.10       & 39.30 & 38.95 \\
    \midrule
    avg & 31.13 & 25.16 & 37.00 & 37.00 & \textbf{38.65} & 33.29 & 37.90 & 38.02 \\
    \bottomrule
    \end{tabular}
    }
    \caption{F1 score in SDQC task of RumourEval-2019 \cite{rumour:19}\label{tab:SDQC}}
\end{table*}

\section{Analysis and Discussion} \label{sec:discussion}
We have shown that sequential subspace alignment is useful across natural language processing tasks. %, however with differences in results between tasks.
For the PeerRead data set we were particularly successful. This might be explained by the fact that the topic of a paper is a simple feature for SSA to pick up on, while being predictive of a paper's acceptance chances.
For NER, on the other hand, named entities can change in less predictable ways across time, proving a larger challenge for our approach.
For SDQC, we were successful in cases where the tweets are nicely clustered by class. For instance, where both rumours are about terrorist attacks, many of the \texttt{support} tweets were headlines from reputable newspaper agencies. These agencies structure tweets in a way that is consistently dissimilar from \texttt{comments} and \texttt{queries}.

The effect of our unbounded time extension boosts results on the PeerRead data set, as the data stretches across a range of years.
In the case of NER, however, this extension is excessive as only two time steps are available.
In the case of SDQC, the lack of improvement could be due to the irregular time intervals, making it hard to learn consistent mappings from rumour to rumour.
Adding instance similarity clustering aids alignment, since considering sample similarities across classes is important over longer time scales.

\subsection{Example of Aligning Tweets}
Finally, we set up the following simplified experiment to investigate the effect of alignment on SDQC data. First, we consider the rumour \texttt{charliehebdo}, where we picked the following tweet:

\vspace{5px}
\noindent \fbox{
	\parbox{.45\textwidth}{%
		\small
		\textbf{Support:}\\ 
		\texttt{France: 10 people dead after shooting at HQ of satirical weekly newspaper \#CharlieHebdo, according to witnesses <URL>}
	}
}
\vspace{5px}

\noindent It has been labeled to be in support of the veracity of the rumour. We will consider the scenario where we use this tweet and others involving the \texttt{charliehebdo} incident to predict author stance in the rumour \texttt{germanwings-crash}. Before alignment, the following 2 \texttt{germanwings-crash} tweets are among the nearest neighbours in the embedding space:

\vspace{5px}
\noindent \fbox{
	\parbox{.45\textwidth}{%
    \small
   \textbf{Query:} \\
   \texttt{@USER @USER if they had, it’s likely the descent rate would’ve been steeper and the speed not reduce, no ?}
	}
}
\vspace{5px}
\noindent \fbox{
	\parbox{.45\textwidth}{%
    \small
    \textbf{Comment:} \\
   \texttt{@USER Praying for the families and friends of those involved in crash. I'm so sorry for your loss.}
}}

%\vspace{5px}
%\noindent \fbox{
%	\parbox{.45\textwidth}{%
%    \small
%    Comment: %\\
%    \texttt{@USER His Facebook, <URL> Address coming shortly}
%}}
%
%\vspace{5px}
%\noindent \fbox{
%	\parbox{.45\textwidth}{%
%     \small
%     @AirSmolik descent rate is not that unusual. 	
%% \end{tcolorbox}

\noindent The second tweet is semantically similar (both are on the topic of tragedy), but the other is unrelated. Note that the news agency tweet differs from the comment and query tweets in that it stems from a reputable source, mentions details and includes a reference. After alignment, the \texttt{charliehebdo} tweet has the following 2 nearest neighbours:

\vspace{5px}
\noindent \fbox{
	\parbox{.45\textwidth}{%
    \small
    \textbf{Support:} \\
    \texttt{“@USER: 148 passengers were on board \#GermanWings Airbus A320 which has crashed in the southern French Alps <URL>”}
}}
%\begin{tcolorbox}[colframe=gray,colback=white,coltext=blue,boxsep=2pt,left=2pt,right=2pt,top=2pt,bottom=2pt]
%    \small
%    Comment: %\\
%    \texttt{@USER @USER radio call then passengers, no ? 20 sec isn’t much… it takes me longer to tweet sometimes ;-)}
%\end{tcolorbox}
\vspace{5px}
\noindent \fbox{
	\parbox{.45\textwidth}{%
    \small
    \textbf{Support:} \\
    \texttt{Report: Co-Pilot Locked Out Of Cockpit Before Fatal Plane Crash <URL> \#Germanwings <URL>}
}}
%% \begin{tcolorbox}[colframe=gray,colback=white,coltext=blue]
%%     \small
%%     @Taxpayers1234 @wingedcrossbill @gatewaypundit This is what happens when society shuns religion. Islam is filling a void!! \#WakeUpAmerica 	
%% \end{tcolorbox}
Now, both neighbours are of the \texttt{support} class. This example shows that semi-supervised alignment maps source tweets from one class close to target tweets of the same class. 

\subsection{Limitations}
A necessary assumption in subspace alignment is that classes are clustered in the embedding space: most embedded tokens should lie closer to \textit{other} embedded tokens of the \textit{same} class than to embedded tokens of another class. If this is not the case, then aligning based on a few labelled samples of class $k$ does not imply that the embedded source tokens are aligned to other target points of class $k$. This assumption is violated if, for instance, people only discuss one aspect of a rumour on day one and discuss several aspects of a rumour simultaneously on day two. One would observe a single cluster of token embeddings for supporters of the rumour initially and several clusters at a later time-step. 
Note that there is no unique solution for aligning a single cluster to multiple clusters.

Additionally, if those few samples labeled in the current time-step (for semi-supervised alignment) are falsely labeled or their label is ambiguous (e.g. a tweet that could equally be labeled as {\sc query} or {\sc deny}), then the source data could be aligned to the wrong point cloud. It is important that the few labeled tokens actually represent their classes. This is a common requirement in semi-supervised learning and is not specific to sequential alignment of text representations.

\subsection{Related Work}

% \paragraph{Temporal Effects in NLP.} 
The temporal nature of data can have a significant impact in natural language processing tasks. % temporal sequences of domains 
For instance, \citeauthor{kutuzov2018diachronic} compare a number of approaches to diachronic word embeddings, and detection of semantic shifts across time.
For instance, such representations can be used to uncover changes of word meanings, or senses of new words altogether \cite{gulordava2011distributional,heyer2009change,michel2011quantitative,mitra2014senses,wijaya2011understanding}.
Other work has investigated changes in the usage of parts of speech across time \cite{mihalcea2012word}.
\citeauthor{yao2018dynamic} investigate the changing meanings and associations of words across time, in the perspective of language change. By learning time-aware embeddings, they are able to outperform standard word representation learning algorithms, and can discover, e.g., equivalent technologies through time.
\citeauthor{lukes:2018} show that lexical features can change their polarity across time, which can have a significant impact in sentiment analysis.
\citeauthor{wang2006topics} show that associating topics with continuous distributions of timestamps yields substantial improvements in terms of topic prediction and interpretation of trends.
Temporal effects in NLP have also been studied in the context of scientific journals, for instance in the context of emerging themes and viewpoints \cite{blei2006dynamic,sipos2012temporal}, and in terms of topic modelling on news corpora across time \cite{allan2001temporal}.
Finally, in the context of rumour stance classification, \citeauthor{lukasik:2016} show that temporal information as a feature in addition to textual content offers an improvement in results.
While this previous work has highlighted the extent to which language change across time is relevant for NLP, we present a concrete approach to taking advantage of this change. Nonetheless, these results could inspire more specialised forms of sequential adaptation for specific tasks.

% \subsection{Temporal Domain Adaptation}
Unsupervised subspace alignment has been used in computer vision to adapt between various types of representations of objects, such as high-definition photos, online retail images and illustrations \cite{fernando2013unsupervised}.%,zhang2017joint}. 
% One first extracts a vector basis in each domain, e.g.~principal components. The basis in the source domain is rotated and translated to align with the basis in the target domain. As both sets of bases contain domain-specific noise components, alignment is based on a subspace. 
Alignment is not restricted to linear transformations, but can be made non-linear through kernelisation \cite{aljundi2015landmarks}. %As stated, \textit{unsupervised} subspace alignment merely aligns the total data structure in each domain and cannot handle class-specific changes between domains. 
An extension to semi-supervised alignment has been done for images \cite{yao2015semi}, but not in the context of classification of text embeddings or domain adaptation on a sequential basis. %In general, domain adaptation in dynamical learning settings has not been done before, to the best of our knowledge.
%Nonetheless, something similar has been proposed to evaluate the quality of word embeddings, where the embedding is aligned to features extracted from manually constructed lexical resources \cite{tsvetkov2015evaluation}. Note that this is not about classification.
% It is also reminiscent of the approach taken by \citeauthor{lample:2018} for aligning languages in unsupervised machine translation.

\section{Conclusions}
%We investigated whether access to future data can be used to adapt data from the past. We evaluated this type of temporal domain adaptation on three different tasks, and showed that using only a handful of data points can yield substantial improvements.

In this paper, we introduced sequential subspace alignment (SSA) for natural language processing (NLP), which allows for improved generalisation from past to present data. Experimental evidence shows that this method is useful across diverse NLP tasks, in various temporal settings ranging from weeks to years, and for word-level and document-level representations. The best-performing SSA method, aligning sub-spaces in a semi-supervised way, outperforms simply training on all data with no alignment.

\section*{Acknowledgements}
WMK was supported by the Niels Stensen Fellowship. 
% Finnish cluster, nvidia, ... 

\bibliographystyle{aaai}
\bibliography{bttf_aaai20}

\end{document}